\documentclass{article}
\usepackage{spconf,amsmath,graphicx}

\usepackage{bm}
\usepackage{multirow}
\usepackage{multicol}
\usepackage{caption}
\usepackage{subcaption}
\usepackage{booktabs}

\usepackage{xcolor}
\usepackage{algorithm}
\usepackage{algpseudocode}

\title{SpeechDPR: End-to-End Spoken Passage Retrieval for Open-Domain Spoken Question Answering}
\name{
\begin{tabular}{@{}c@{}}
Chyi-Jiunn Lin$^1$, Guan-Ting Lin$^1$, Yung-Sung Chuang$^2$, Wei-Lun Wu$^1$, Shang-Wen Li$^3$, \\Abdelrahman Mohamed$^4$, Hung-yi Lee$^1$, Lin-shan Lee$^1$
\end{tabular}
}
\address{
$^1$National Taiwan University \quad
$^2$Massachusetts Institute of Technology \\
$^3$Meta AI \quad
$^4$Rembrand 
}
\begin{document}
\ninept
\maketitle
\begin{abstract}

Spoken Question Answering (SQA) is essential for machines to reply to user's question by finding the answer span within a given spoken passage. 
SQA has been previously achieved without ASR to avoid recognition errors and Out-of-Vocabulary (OOV) problems. 
However, the real-world problem of Open-domain SQA (openSQA), in which the machine needs to first retrieve passages that possibly contain the answer from a spoken archive in addition, was never considered. This paper proposes the first known end-to-end framework, Speech Dense Passage Retriever (SpeechDPR), for the retrieval component of the openSQA problem. 
SpeechDPR learns a sentence-level semantic representation by distilling knowledge from the cascading model of unsupervised ASR (UASR) and text dense retriever (TDR).  
No manually transcribed speech data is needed. 
Initial experiments showed performance comparable to the cascading model of UASR and TDR, and significantly better when UASR was poor, verifying this approach is more robust to speech recognition errors.

\end{abstract}
\section{Introduction}
\label{sec:intro}

Spoken Question Answering (SQA) is to find the answer span out of a passage for a given question by a machine, where the question, passage and answer span are all in the form of audio waveform (the question may be in text form in some cases though) \cite{turmo2008overview, comas2012sibyl, lee2018spoken, lee2018odsqa}. Open-domain SQA (openSQA) is one step further, in which the passage is not given, and the machine has to find in addition one or more passages containing the answer from a large spoken dataset or the spoken content over the Internet, referred to as ``a spoken archive" below, before performing SQA. These technologies are certainly important when machines have to reply to the user's questions.

%SQA is a difficult task compared to many other speech understanding tasks like translation or intent classification, because the required understanding is on the utterance level for the latter but over much longer passages for the former. Furthermore, for SQA, the machine is required to comprehend the overall knowledge offered by the entire passage and to catch the fine-grained information to locate the answer span precisely. OpenSQA is a more realistic task for the real-world scenario in which the spoken passages for the question are not given. OpenSQA is a much more challenging task than SQA alone because retrieving passages from a spoken archive semantically relevant to the user's question is difficult.

Substantial effort has been made and reasonable results have been obtained for the text version of the above two tasks, Text Question Answering (TQA) and Open-domain TQA (openTQA), in which everything is in form of text instead. The latter is usually achieved by cascading a text retriever in front of a text reader (or TQA) \cite{chen2017reading, yang2019end, karpukhin2020dense}. So a natural approach for SQA and openSQA discussed here is simply cascading an ASR module in front of TQA or openTQA, or performing TQA or openTQA on top of the ASR transcriptions.

%The major problem for the above cascading approach is the inevitable recognition errors produced by the ASR module can not be recovered by the following TQA or openTQA, but instead cause more errors. Moreover, the correct answers to the questions very often include named entities or out-of-vocabulary (OOV) words which can never be recognized. Also, the ASR module was trained by minimizing WER in which function words irrelevant to SQA task are equally counted as keywords, so the cascade of two modules (ASR and TQA) individually optimized with two different criteria can never perform as well as a single model globally optimized. Not to mention reliable ASR modules have to be trained with large quantities of labeled data, which are prohibitively difficult for the many low-resourced languages over the world. Although this issue can be addressed by training unsupervised ASR (UASR) \cite{baevski2021unsupervised, liu2023towards} with unpaired speech and text data, current unsupervised ASR models are still less performant and stable than the supervised ones. These explain why end-to-end approaches directly identifying the answer span over the audio signals rather than over the ASR transcriptions, eliminating the need of an ASR module, are highly desired.

%\textbf{
However, building reliable ASR modules requires extensive training with large quantities of labeled data, which is prohibitively difficult for many low-resourced languages over the world.
To render openSQA affordable for the low-resourced languages, it is imperative to develop models that require no paired speech-text data for both training and inference. 
Therefore, this paper specifically emphasizes on the scenario in which \textit{no paired speech-text data is available}.
%}

%\textbf{
A simple solution is to replace the supervised ASR with an unsupervised ASR (UASR) \cite{baevski2021unsupervised, liu2023towards} trained on unpaired speech and text data.
However, current unsupervised ASR models are still less performant and stable than the supervised ones.
Besides, the recognition errors produced by the inaccurate UASR module can not be recovered by the following TQA or openTQA, but instead cause more errors.
Moreover, the correct answers to the questions often include named entities or out-of-vocabulary (OOV) words which can never be recognized.
These explain why end-to-end approaches directly performing retrieval and identifying the answer span over the audio signals rather than over the UASR transcriptions are highly desired.
%}

%\textbf{[TODO: carify that SpeechBERT and SPLAT still require ASR] }
SpeechBERT \cite{chuang2020speechbert}  performed end-to-end SQA by properly aligning the phonetic embeddings of spoken words and semantic embeddings of text words in a hidden space, so the semantics can be extracted directly from the audio signals. SPLAT \cite{chung2021splat} achieved similar goals later on with improved performance. DUAL \cite{2022arXiv220304911L} was then the first successfully performing SQA without any paired speech-text data by transforming the audio signals into sequences of quantized discrete units such that textless NLP \cite{lakhotia2021generative, lee2021direct} can be performed bypassing ASR errors. 
However, all of these above mentioned approaches achieved SQA only, not openSQA. 
In these approaches, the answer spans were identified from given passages, while the problem of how passages containing the answer to the user's question can be obtained from a large spoken archive was totally left out.  

%\textbf{
While the problem of retrieving passages from spoken archives has been extensively investigated and reported \cite{lee2015spoken, garofolo2000trec, larson2012spoken,chelba2008retrieval, witbrock1997speech}, prior works have not specifically addressed the end-to-end resolution of the openSQA retrieval task. The openSQA retrieval task involves semantic retrieval from sentence to sentence, where the question is a spoken sentence that may not necessarily share overlapping content with the gold passage.
Instead, previous end-to-end approaches were limited to either query-by-example spoken-term detection \cite{ram2020neural, ao2018query}, where the question (or query) typically consists of a spoken short phrase and the gold passages must contain this phrase, or semantic search limited to visually-grounded speech \cite{harwath2018vision, kamper2018semantic}.
%The openSQA retrieval task, however, is a sentence-to-sentence semantic retrieval problem where the question is a sentence not necessarily having overlapping content with the gold passage, and no paired visual grounded data is available.
%}

\begin{figure}[t]
  \centering
  \includegraphics[width=0.95\linewidth]{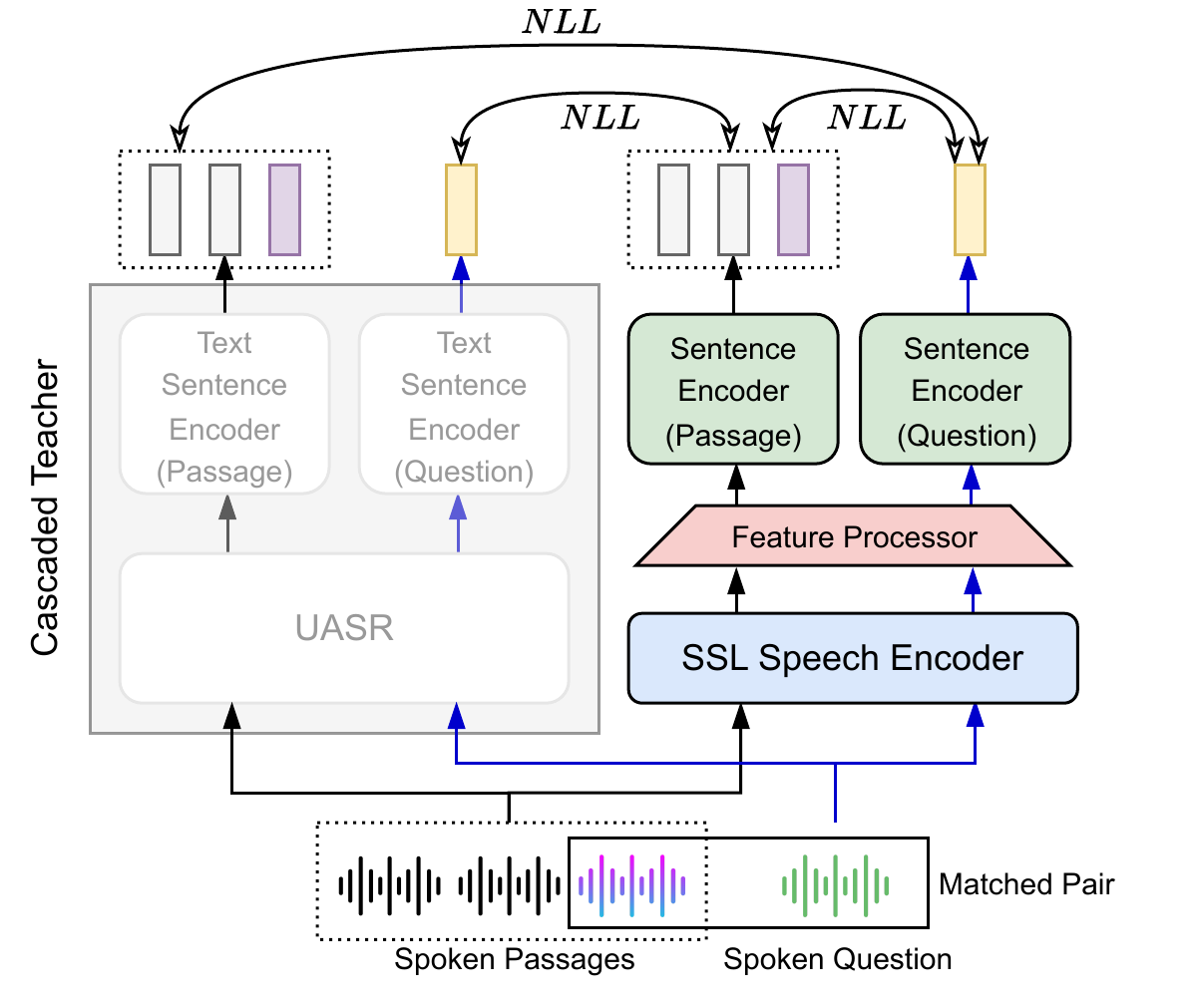}
    \vspace{-0.5cm}
  \caption{The framework of SpeechDPR.}
 \vspace{-0.5cm}
  \label{fig:scenario}
\end{figure}

%\textcolor{red}{
%On the other hand, for the cascading approach, although Sidiropoulos et al. \cite{sidiropoulos2022impact} augmented the question text with synthetic ASR noise to mitigate the effect of question ASR noise on text passage retrieval, the augmentation relied on the text-to-speech model trained with paired speech and text data, so this method may not generalize to the scenario of no transcribed speech data.
 %%Besides, they conducted experiments on the text passage without ASR noise and did not investigate whether the method also worked well in retrieving ASR-transcribed text passages. 
 %}

This paper addresses the untranscribed spoken-question-to-spoken-passage semantic retrieval problem for openSQA by proposing an end-to-end model, SpeechDPR (Speech Dense Passage Retriever), to retrieve the passages from a spoken archive without supervised ASR or manually transcribed speech data.
SpeechDPR adopts the bi-encoder retriever framework and learns a sentence-level semantic representation space by distilling knowledge from the cascading model of UASR and text dense retriever (TDR). 
SpeechDPR assesses the similarity between the question and each passage in the spoken archive by calculating the inner product of their sentence representations, and thus can find the most semantically relevant passage.

% brief summary
The main contributions are summarized as follows: 
\begin{itemize}
\item This research proposes SpeechDPR, the first end-to-end model to tackle the challenge of untranscribed spoken passage semantic retrieval for openSQA without any supervised ASR transcriptions or speech-text paired data for training and inference.
\item SpeechDPR achieves a competitive retrieval accuracy, comparable to the cascading baselines, which involves performing the corresponding modules trained on text on top of UASR transcriptions, and outperforms significantly in scenarios with relatively poor ASR accuracies.
\end{itemize}

\section{Background}
%\subsubsection{Problem formulation}
We propose the never-attempted retrieval-SQA framework to address the openSQA problem, and this paper focuses on improving the retrieval component of openSQA, where all questions and passages are in the form of raw spoken sentence waveforms.
Given a spoken question and a spoken archive containing numerous long spoken passages, such as broadcast news or podcast database, the retrieval model calculates the similarity between each passage and the question. 
It then returns a small number of passages with the highest similarity so that the downstream SQA model can identify the answer span from these passages.
A retrieval is considered successful when the returned passages includes the gold passage containing the answer to the question.
We assume neither paired speech-text data nor off-the-shelf supervised ASR is available.

\subsection{Bi-encoder dense retriever framework}
%\textbf{TO REVISE}

SpeechDPR and all TDRs in this paper adopt the bi-encoder dense retriever framework commonly used in openTQA~\cite{karpukhin2020dense, xiong2020approximate}.
The bi-encoder dense retriever encodes the question and the passage with two separate encoders.
Given a bi-encoder retriever model $(Q, P)$ with a question encoder $Q(.)$ and a passage encoder $P(.)$, before performing retrieval, the retrieval model encodes each passage $\bm{p}$ in the archive into a low-dimensional continuous vector $P(\bm{p})$. 
During retrieval, the model similarly encodes the input question $\bm{q}$ into a vector $Q(\bm{q})$ so that the similarity between $\bm{q}$ and each $\bm{p}$ can be expressed by calculating the dot product of $Q(\bm{q})$ and $P(\bm{p})$ \cite{karpukhin2020dense}:
\begin{equation}
sim(\bm{q}, \bm{p}) = Q(\bm{q}) \cdot P(\bm{p}) 
\end{equation}

 The bi-encoder dense retriever (Q, P) learns a sentence representation that minimize the negative log likelihood (NLL) between the representation of the question and its paired passage. 
For each pair of question and its relevant (positive) passage $(\bm{q}, \bm{p}^{+})$ in the dataset, $n$ irrelevant (negative) passages $\{\bm{p}^{-}_{i}\}^{n}_{i=1}$ 
 are sampled to calculate the NLL loss:
\begin{equation}
\begin{aligned}
\label{eq:nll}
 NLL_{Q, P}  = - log \frac{e^{Q(\bm{q}) \cdot P(\bm{p}^{+})}}{e^{Q(\bm{q}) \cdot P(\bm{p}^{+})} +  \sum_{i=1}^{n} e^{Q(\bm{q}) \cdot P(\bm{p}^{-}_{i})}}
\end{aligned}
\end{equation}

In this paper, the negative passages we use to calculate the NLL loss are the gold passages of the other questions in the same batch.

\section{Proposed approach}
\label{sec:proposed-approach}

\subsection{SpeechDPR model}
%\subsubsection{Model architecture}
%As shown in Figure~\ref{TODO}, 
The end-to-end SpeechDPR model is composed of SSL speech encoder, feature processor, question sentence encoder, and passage sentence encoder. In forward propagation, the input waveform is passed through the SSL speech encoder, and downsampled by feature processor. Afterwards, the question or passage encoder is used to extract semantic embedding according to the type of the speech. 

SSL speech encoder is pre-trained on raw speech waveforms via self-supervised learning (SSL) and turns the waveform into contextualized frame-level representations, or a sequence of vectors. The SSL encoder is only used for feature extraction, and its parameters are frozen in our experiments.
% (1) the version using CNN layers
Feature processor instance-normalizes the extracted representations and passes them through a two-layer CNN to shorten the very long sequence length for reducing memory consumption. 
% (2) the version using ASR word segments
%In Feature processor, the extracted representations are instance normalized and then segmented according to UASR word segments. 
%Afterward, we apply weighted pooling, in which the weights are determined by a two-layer MLP, on the segments of representations to get a sequence of segment-level representations.
%Finally, the representations are passed through a linear projection layer to fit the hidden dimension of Sentence encoder.
Both sentence encoders use the Roberta-base~\cite{liu2019roberta} encoder to encode the processed speech representations.
This design choice is based on the finding of BERT-like encoders' cross-disciplinary transferability by Kao et al.~\cite{kao2021bert}.
The representation sequence is first concatenated with Roberta word embedding of the [CLS] token at the beginning and fed into the sentence encoder to derive the sentence-level representation, a 768-dim vector, from the representation of the [CLS] token. 

% We freeze the parameters of the SSL speech encoder, training other components jointly. 
% All of the above-mentioned modules except SSL speech encoder are trained during SpeechDPR training.

%\subsection{Proposed approach: SpeechDPR}
\subsection{Knowledge distillation from Cascading Teacher model}
%\textbf{[TODO: motivation, (preliminary experiments show that SpeechDPR requires knowledge distillation to warm-start training)]}
Based on our preliminary finding that SpeechDPR performs incompetently if it is trained by just minimizing the NLL loss in Eq. \ref{eq:nll}, we additionally distill knowledge from Cascading Teacher model. The Cascading Teacher consists of a UASR and a TDR, neither of which requires paired speech-text data for training in our case. Specifically, UASR transcribes speech waveforms to text sentences and is trained on unpaired speech-text data via adversarial learning.
TDR is a bi-encoder text retriever that converts the UASR-transcribed text sentence to the sentence-level representation. 
In this paper, we train the TDR with UASR transcripts instead of the ground truth transcripts. 

\subsection{Training objective}
We propose to distill the sentence representation encoded by TDR to improve SpeechDPR training.
Apart from learning to minimize the NLL loss in Eq. \ref{eq:nll}, SpeechDPR additionally distills knowledge from Cascading Teacher model by minimizing the NLL between the question sentence representation encoded by SpeechDPR and the passage sentence representation encoded by the teacher, and similarly minimizing the NLL between the passage sentence representation encoded by SpeechDPR and the question sentence representation encoded by the teacher.
Therefore, we set the total loss to the weighted sum of the above-mentioned three NLL losses.
Let $Q_T, P_T$ respectively denote the question and passage encoder of the teacher model, and $Q_S, P_S$ denote those of the student model, which is SpeechDPR here.
%For each pair of question and its relevant (positive) passage $(\bm{q}, \bm{p}^{+})$ in the dataset, we sample $n$ irrelevant (negative) passage $\{\bm{p}^{-}_{i}\}^{n}_{i=1}$ to calculate the NLL loss.
The total loss function $L$ is:
\begin{equation}
\label{eq:loss-function}
L = NLL_{Q_S, P_S} + \alpha NLL_{Q_S, P_T} + \beta NLL_{Q_T, P_S},
\end{equation}
where $\alpha$ and $\beta$ are tuned hyperparameters.

\section{Experiments}
\label{sec:experiments}
% 1. corpus
% 2. evaluation (retrieval ; QA : speechbert-type / MFA-NLP-type)
% 3. cascade approach (intro to 2 ASR module)
\subsection{Data}
\label{ssec:data}
We adopt a data setup similar to that used in openTQA research~\cite{chen2017reading, karpukhin2020dense}.
These studies gathered questions from various TQA datasets and used a set of Wikipedia passages as the sole source for retrieval during both training and inference.
In our paper, we use spoken questions from the SLUE-SQA-5 dataset \cite{shon2022slue} and spoken passages from the Spoken Wikipedia dataset \cite{baumann2019spoken}.

SLUE-SQA-5 is a SQA dataset whose training, dev, test subsets all includes the human-spoken version of questions sourced from five TQA datasets: TriviaQA~\cite{joshi2017triviaqa}, SQuAD v1.1~\cite{rajpurkar2016squad}, Natural Questions (NQ)~\cite{kwiatkowski2019natural}, WebQuestions (WQ)~\cite{berant2013semantic} and CuratedTREC (TREC)~\cite{baudivs2015modeling}. 
Notably, the question speakers across different subsets are distinct.
%, resulting in a total of 15 disjoint sets of speakers.
This characteristic makes SLUE-SQA-5 well-suited for openSQA experiments.
The dataset contains about 46k questions in the training set, 1.9k in the dev set, and 2.4k in the test set.
Each paired passage of SLUE-SQA-5 questions consists of a 40-second Spoken Wikipedia passage that includes the answer to the corresponding question. 

Spoken Wikipedia contains the spoken version of 1.2k Wikipedia articles from about 400 human speakers.
This paper follows the pre-processing procedure in SLUE-SQA-5 by splitting spoken articles into about 39k spoken passages with duration of 40 seconds and taking these passages as spoken archive for retrieval.
The total duration of the spoken archive is 427 hours.

Note that currently due to a lack of suitable datasets for openSQA in other languages, we are limited to experimenting on English.

\subsection{Evaluation}
%\subsection{Evaluation of retrieval}
Top-K retrieval accuracy, or the percentage of questions for which the top-K passages returned by the retriever include any gold passage, is used to evaluate the retriever performance. 
%Unlike text version of retrieval, here we assume the transcriptions of the spoken passages are not available, so we can consider a passage as the gold passage to a question only when it is paired with that question in the original SLUE-SQA-5 setting.
We consider a passage as the gold passage to a question only when it is paired with that question in the original SLUE-SQA-5 setting.
This paper reports the top-20 accuracy.

%\subsection{Evaluation of openSQA}
%% the frame level f1, AOS
%\textbf{(TODO: add reader details and how the final answer span is derived, using top-20 passages.)} 
In addition, to investigate how different retrievers affect the whole openSQA accuracy, we passed the top-20 retrieved passages into a shared SQA module to predict the answer span to the question.
The implementation details of the SQA module are described in Section \ref{ssection:implementation}.
Following the works \cite{chuang2020speechbert, shon2022slue} in SQA, we use the FF1 (Frame-level F1 Score, overlapping span length out of the reference span and predicted span respectively as precision and recall) as the metrics for evaluating the performance of openSQA.
%% details about FF1 and AOS
%Both of them measure the ratio of overlapping between the predicted answer span and gold answer span at time level.
Note that sometimes the correct answer is in the answer span identified by SQA, but because the carrying passage is somehow not labeled as a gold passage for the question, and we assume the transcriptions are not available, this answer span can only be taken as incorrect and the FF1 is set to zero. This may underestimate the performance of the models.

\subsection{Implementation details}
\label{ssection:implementation}

% cascading teacher model
% UASR

\textbf{Cascading Teacher model.} 
%We follow the training procedure of wav2vec-U 2.0~\cite{liu2023towards} to train our UASR module on the datasets mentioned in Section~\ref{ssec:data}.
%We pick the unpaired speech data in SLUE-SQA-5 training set and the unpaired text data in TriviaQA training set excluding the ones included in SLUE-SQA-5 for the adversarial-training stage, while all spoken passages in Spoken Wikipedia plus the SLUE-SQA-5 training set are used for the self-training stage.
%The UASR model has word error rates of 24.1, 23.3, 39.8 respectively on SLUE-SQA-5 dev set, SLUE-SQA-5 test set, Spoken Wikipedia passages.
For the UASR module, we adopt the training procedure of wav2vec-U 2.0~\cite{liu2023towards}.
In the adversarial-training stage, we use the unpaired speech data in SLUE-SQA-5 training set and the unpaired text data in TriviaQA training set excluding the ones included in SLUE-SQA-5.
In the self-training stage, we use speech data in Spoken Wikipedia and the SLUE-SQA-5 training set.
% TDR
For the TDR module, we use two Roberta-base models as the bi-encoder and train the models on the UASR transcriptions with 64 batch size, 4e-5 learning rate, and 100 warmup steps for 100 epochs.

% speechDPR
\textbf{SpeechDPR.} We use the HuBERT-large model\footnote{https://huggingface.co/facebook/hubert-large-ll60k} with 24 transformer layers pre-trained on Libri-Light speech dataset~\cite{kahn2020libri} as SSL speech encoder and extract representations from its 22nd layer.
The CNN in feature processor has a stride of 4 and 3 in its first and second layer, respectively.
Feature processor and the two sentence encoders are jointly trained with 64 batch size, 1e-4 learning rate, and 500 warmup steps for 100 epochs.
%The word boundaries are derived by aligning the UASR transcripts to speech with Montreal Forced Aligner  \cite{mcauliffe2017montreal}.
We set both $\alpha$ and $\beta$ to 0.5 in the loss function (Eq.~\ref{eq:loss-function}).

The above-mentioned hyperparameters and the best checkpoints of both TDR and SpeechDPR models are picked according to the top-20 retrieval accuracy on SLUE-SQA-5 dev set.
%\textbf{[TODO: mention that weight of ensembling are tuned according to accuracy on dev set, too.]}

\textbf{The shared SQA module for openSQA evaluation.}
The SQA module is a cascading model of UASR mentioned above and a Deberta-base \cite{he2020deberta} TQA model fine-tuned on SQuAD v1.1 questions excluding the ones included in the SLUE-SQA-5.
During openSQA inference, we pick the answer span with highest answer score from the retrieved 20 passages as the finally predicted answer span.
Following \cite{yang2019end}, the answer score is defined as the linear interpolation of the passage similarity score predicted by the retriever and the span score predicted by the SQA module.
The weights for linear interpolation are tuned on the SLUE-SQA-5 dev set.
To evaluate FF1, we use Montreal Forced Aligner \cite{mcauliffe2017montreal} to obtain the time intervals of the TQA-predicted answers in seconds.
When given the corresponding gold passage to each question, the SQA module obtains 11.17 FF1 on SLUE-SQA-5 test set.

\subsection{Baseline: the cascading approach (UASR + TDR)}
Because there did not exist any known prior work for the openSQA retrieval task considered here, we take the cascading approach (UASR + TDR) as the baseline to compare with. 
Apart from Cascading Teacher described above, for a fairer comparison, we additionally train another cascading model, referred to as "Cascading Student", by distilling knowledge from Cascading Teacher.
Here we apply the same objective function with that of SpeechDPR (Eq.~\ref{eq:loss-function}).
Cascading Student shares the same UASR module and hyperparameters of TDR training with the teacher.

\section{Results}
\label{sec:results}
\iffalse
%
\begin{table*}[!h]
  \caption{Top-20 retrieval accuracy (\%) (percentage of questions whose top-20 retrieved passages include at least one gold passage) and end-to-end openSQA accuracy (FF1) on SLUE-SQA-5 test set questions. To show domain generalizability, we also report the accuracies on the TriviaQA-split questions (\textbf{TriviaQA}) and other questions (\textbf{Others}) in the test set apart from the accuracies on the whole test set (\textbf{Overall}).}
  \label{tab:retrieval-accuracy}
  \centering
  \begin{tabular}{l|l|ccc|ccc}
    \toprule
    \multirow{2}{*}{\textbf{Training Questions}} & \multirow{2}{*}{\textbf{Retrieval Model}} & \multicolumn{3}{c} {\textbf{Top-20 retrieval accuracy}}  & \multicolumn{3}{c}{\textbf{FF1}} \\
    & & \textbf{TriviaQA} & \textbf{Others} & \textbf{Overall} & \textbf{TriviaQA} & \textbf{Others} & \textbf{Overall}  \\
     \midrule
     & (a) Baseline: Cascading Teacher &  & & & & &  \\
      & (b) Baseline: Cascading Student & & & & & & \\
     Whole SLUE-SQA-5  & (c) Proposed: SpeechDPR & & & & & & \\
      \cmidrule(lr){2-8}
       training set & Ensemble of (a) and (b) & & & & & &  \\
      & Ensemble of (a) and (c) & & & & & & \\
      \midrule
     & (a) Baseline: Cascading Teacher &  & & & & & \\
      & (b) Baseline: Cascading Student & & & & & & \\
     Only TriviaQA-split of  & (c) Proposed: SpeechDPR & & & & & &  \\
      \cmidrule(lr){2-8}
     SLUE-SQA-5 training set & Ensemble of (a) and (b) & & & & & & \\
      & Ensemble of (a) and (c) & & & & & &  \\

    \bottomrule
  \end{tabular}
\end{table*}
%
\fi

\subsection{Retrieval results}
\label{ssec:retr-result}
\begin{table}[!t]
  \caption{Top-20 retrieval accuracy (\%) (Top-20) and openSQA accuracy (FF1) on SLUE-SQA-5 test set questions. "Knowledge distillation from Cascading Teacher" is abbreviated as "KD" in row (d). The openSQA accuracy is derived by passing the top-20 retrieved passages to a shared SQA model to evaluate FF1.}
  \vspace{-0.3cm}
  \label{tab:retrieval-accuracy}
  \centering
  %\resizebox{0.75\columnwidth}{!}{
  \begin{tabular}{l|c|c}
    \toprule
     \textbf{Retrieval Model} & \textbf{Top-20}  & \textbf{FF1} \\
    \midrule
    \textbf{Single model:} & & \\
      (a) Baseline: Cascading Teacher & \textbf{19.94} & 0.561  \\
       (b) Baseline: Cascading Student & 19.90 &  \textbf{0.565} \\
      (c) Proposed: SpeechDPR & 19.73 & 0.558 \\
      (d) Ablation: SpeechDPR w/o KD & 00.04 & 0.000 \\
      \midrule
      \textbf{Ensemble model:} & & \\
       Baseline: Ensemble of (a) and (b) & 25.60 &  0.860 \\
       Proposed: Ensemble of (a) and (c) & \textbf{28.88} & \textbf{0.951} \\
    \bottomrule
  \end{tabular}
  %}
  \vspace{-0.3cm}
\end{table}
The top-20 retrieval accuracies are listed in the second column of Table \ref{tab:retrieval-accuracy}. 
By comparing Cascading Student in row (b) to Cascading Teacher in row (a), we see that Cascading Student gets a similar top-20 accuracy (19.90\%) to Cascading Teacher does (19.94\%), indicating that knowledge distillation from text retrievers to text retrievers does not lead to significant improvement.
By comparing SpeechDPR in row (c) to baselines of row (a) and row (b), we observe that SpeechDPR achieves a comparable accuracy (19.73\%) with the two cascading baseline models.
The difference between their retrieval accuracies is less than 1\%.
This result shows that SpeechDPR can extract useful information for retrieval directly from continuous speech waveforms instead of the intermediate textual transcripts while maintaining a reasonable performance.

Besides, we also conduct an ablation study by training another SpeechDPR model without knowledge distillation (KD) from Cascading Teacher, or without the $NLL_{Q_S, P_T}$ and $NLL_{Q_T, P_S}$ terms in the loss function (Eq. \ref{eq:loss-function}).
This trained model, denoted by ``SpeechDPR w/o KD" in row (d), gets a significantly lower accuracy (0.04\%) compared to the original SpeechDPR in row (c).
This indicates that knowledge distillation from a cascading model is crucial for SpeechDPR training.

Although SpeechDPR distills from the Cascading Teacher, it is also trained end-to-end with speech directly. Therefore, SpeechDPR can potentially achieve superior performance on the one performs poorly with Cascading Teacher. The model ensemble technique is a common technique to combine models with different specialty, resulting in better performance. We tried to ensemble the SpeechDPR model in row (c) and with Cascading Teacher in row (a) by linearly interpolating the similarities scored by the two models.
The weights for linear interpolation are tuned according to top-20 retrieval accuracy on the dev set.
This ensemble model gets a retrieval accuracy (28.88\%), which significantly surpasses the accuracy (25.60\%) of the ensemble of Cascading Student in row (b) and Cascading Teacher in row (a). The result verifies that end-to-end SpeechDPR model can learn extra knowledge complementary to that learned by the cascading model, so the ensemble of the two models achieves superior performance to the ensemble of two cascading models.

\subsection{OpenSQA results}
The openSQA accuracies (FF1) are listed in the third column of Table \ref{tab:retrieval-accuracy}. 
We can observe a trend similar to that in the top-20 retrieval accuracy.
The two baseline models in row (a), (b) and SpeechDPR in row (c) achieve similar FF1 scores (0.561, 0.565, 0.558, respectively), and the ensemble of SpeechDPR and Cascading Teacher outperforms the ensemble of the two cascading baselines (0.951 vs. 0.860). 
This result shows the efficacy of SpeechDPR in the whole openSQA task.
Besides, the ablated SpeechDPR in row (d) fails to identify any correct answers obviously because it has a poor retrieval accuracy.

%Note that the openSQA accuracy is bounded by the downstream SQA module performance because SLUE-SQA-5 dataset question is challenging for current SQA models, and even the best-performing model using supervised ASR achieves only 43.3 FF1 score.
It is worth noting that the openSQA accuracy is constrained by the performance of the downstream SQA module. 
This is due to the fact that questions from the SLUE-SQA-5 dataset pose a significant challenge for current SQA models. 
According to the SLUE-SQA-5 paper \cite{shon2022slue}, even the highest-performing model without utilizing paired speech data, achieves only a 21.8 FF1 score on SQA.

\subsection{Retrieval performance at different WERs}
\begin{figure}[t]
    \centering
        \includegraphics[width=0.75\linewidth]{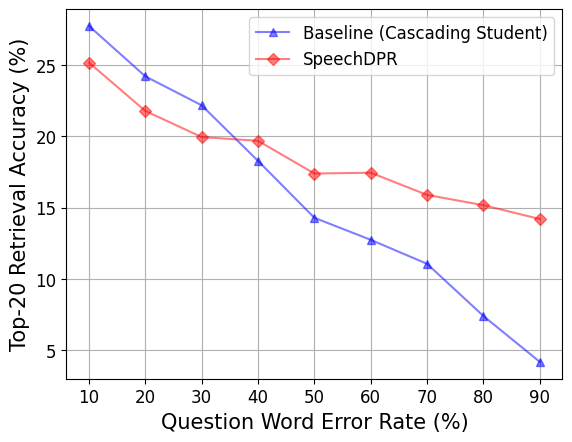}
        \vspace{-0.3cm}
    \caption{Top-20 retrieval accuracy evaluated on subsets of SLUE-SQA-5 test set questions with different levels of UASR WER.}
    \vspace{-0.3cm}
    \label{fig:ff1-top1-wer}
\end{figure}
In real-world scenarios, ASR accuracy can be very poor due to noisy acoustic conditions, out-of-vocabulary (OOV) and other problems, referred to as ASR conditions here. 
We thus investigate the effect of these ASR conditions on the retrieval performance by splitting the questions in SLUE-SQA-5 test set into several subsets according to UASR WER in the question. 
The result of top-20 retrieval accuracy within each of these subsets are plotted respectively in Figure~\ref{fig:ff1-top1-wer}.
% compare results
We can see the retrieval accuracy of the Cascading Student baseline degrades seriously when UASR WER gets higher, while that of the proposed SpeechDPR is more robust to ASR conditions, obviously because SpeechDPR does not rely on UASR for inference.
SpeechDPR actually significantly outperforms the cascading baseline when WER exceeds 40\%.
Because many of the questions and passages in openSQA task include name entities and OOV words, this makes the proposed SpeechDPR highly attractive.

\section{Conclusion}
\label{sec:discussion}
Here we propose SpeechDPR, an end-to-end model for the semantic retrieval task for openSQA by knowledge distillation from the cascading model of UASR and TDR, in which no supervised ASR or any manual transcript is needed.
%, and semantics is directly extracted from the speech waveforms of the spoken question and spoken passage to access their relevance for retrieval.
Experimental results show that SpeechDPR achieves competitive performance compared to the cascading baseline (UASR-TDR) and outperforms significantly in scenarios of poor UASR performance.
This is attractive because many spoken questions and passages include name entities and OOV words which cannot be recognized by UASR.
The initial experiments are only done on English, but our framework can be extended to other low-resourced languages in future research.

\section{Acknowledgements}
We thank National Center for High-performance Computing (NCHC) of
National Applied Research Laboratories (NARLabs) in Taiwan for providing computational and storage resources.
Besides, we thank Yuan Tseng and Chi-Liang Liu in National Taiwan University for useful discussions and suggestions.

\bibliographystyle{IEEEbib}
\bibliography{strings,refs}

\end{document}